\documentclass{article}
\usepackage[T1]{fontenc}
\usepackage{spconf,amsmath,amssymb,graphicx,hyperref}
\usepackage{booktabs,placeins}
\hypersetup{hidelinks}

\title{MHR-Router: Multimodal Heterogeneous Retrieval-Enhanced Adaptive Routing for VLM Inference}

\name{Wenjuan Zhu$^{1,2}$ \qquad Zheming Yang$^{1,2\star}$\thanks{$^{\star}$ Corresponding author.} \qquad Yunqing Hu$^{1,2}$ \qquad Qi Guo$^{1,2}$}
  
\address{
$^{1}$Institute of Computing Technology, Chinese Academy of Sciences, Beijing, China\\
$^{2}$University of Chinese Academy of Sciences, Beijing, China
}

\begin{document}
\ninept

\maketitle

\begin{abstract}
Vision-language models (VLMs) exhibit complementary capabilities, yet selecting the most suitable expert for a given image-question query remains challenging due to heterogeneous multimodal inputs and varying expert strengths. Existing routing approaches rely on text-based cues or static model profiles, failing to capture fine-grained visual evidence and expert-specific failure patterns. To address this, we propose MHR-Router, a multimodal heterogeneous retrieval-enhanced adaptive routing framework for VLM inference. It accurately dispatches each multimodal query to the most suitable VLM expert via local retrieval evidence. MHR-Router consists of two stages: (1) Multimodal Routing Representation, which fuses frozen multimodal features into a comprehensive query representation; and (2) Heterogeneous Retrieval-Enhanced Adaptive Routing, which retrieves expert-specific success and failure memories as local evidence and employs a shared scorer to adaptively select the optimal expert. Experiments show that MHR-Router effectively achieves heterogeneous modality routing and consistently improves accuracy over traditional baselines.

\end{abstract}


\section{Introduction}
\label{sec:intro}

Vision-language models (VLMs) have rapidly evolved into general-purpose interfaces for visual perception and reasoning \cite{gui2026pro222222}. Experts with different architectures, scales, and training data exhibit complementary strengths across object recognition, OCR, spatial reasoning, fine-grained recognition, and anomaly detection\cite{tian2025saec262626}. Because competence varies across tasks and individual inputs, an expert that performs well for one image-question pair may be less suitable for another. A fixed expert therefore leaves complementary capabilities unexploited, whereas evaluating every candidate requires invoking the entire expert pool for each query \cite{zhang2026efficient252525,yang2025moa242424}. VLM routing addresses this trade-off by selecting a single expert for each query. Its central challenge is to estimate expert suitability from the joint image-question evidence before invoking any candidate expert\cite{vsabanovic2026inar272727}.

\begin{figure}[!t]
\centering
\includegraphics[width=\linewidth,trim=20pt 0 0 8pt,clip]{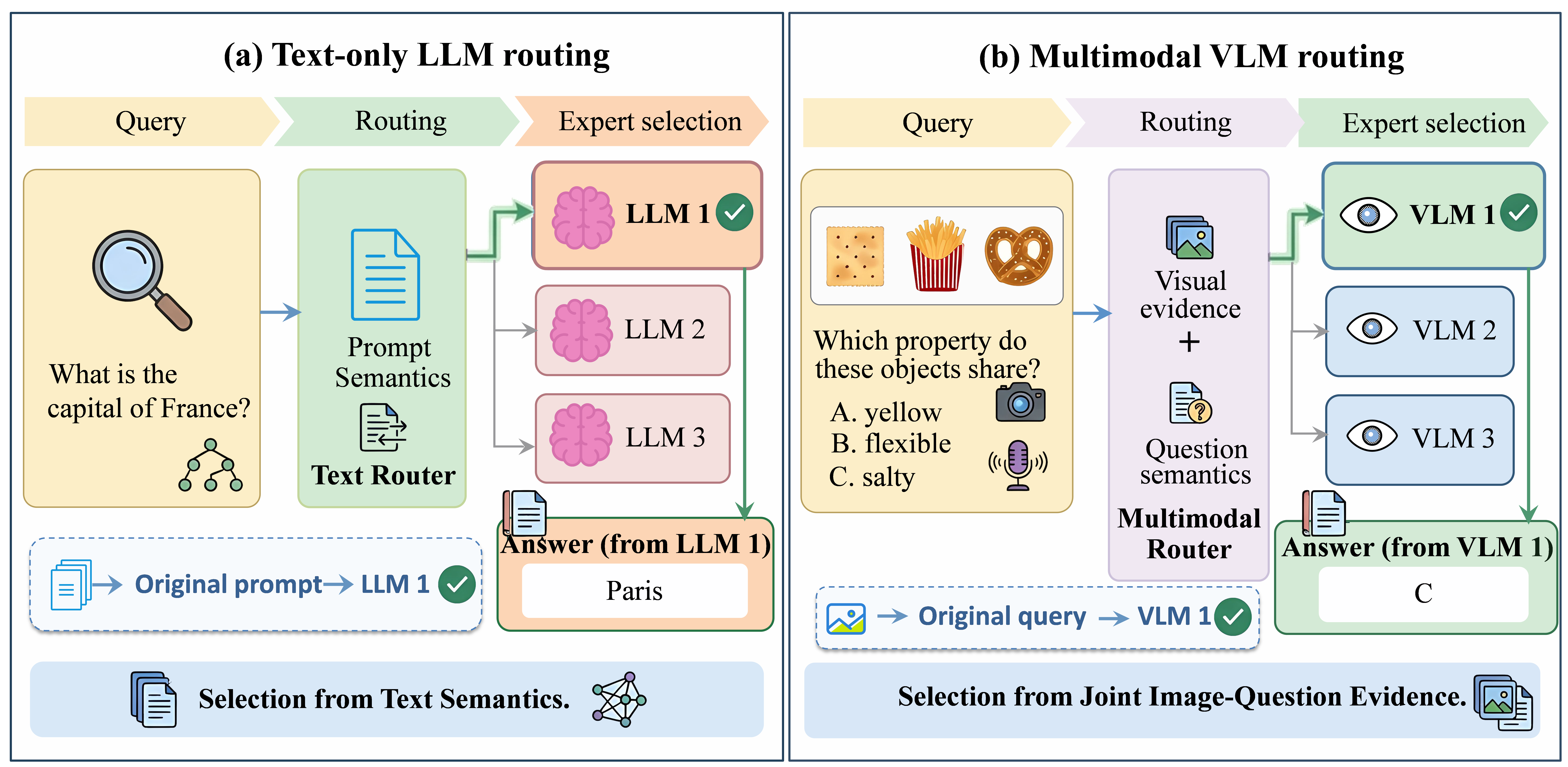}
\caption{Comparison between text-only LLM routing and multimodal VLM routing. (a) A text-only router selects an LLM expert primarily from prompt semantics. (b) A multimodal router selects a VLM expert by jointly considering visual evidence and question intent.}
\label{fig:intro_routing_comparison}
\end{figure}

Most existing model-routing methods, however, were developed for text-only LLMs and estimate expert suitability primarily from prompt semantics or preference data \cite{frugalgpt2023,routellm2024,iclrouter2025}. As illustrated in Fig.~\ref{fig:intro_routing_comparison}, text-only LLM routing selects an expert based on prompt semantics alone, whereas multimodal VLM routing jointly considers visual evidence and question semantics to select an expert for the specific image-question pair. This comparison highlights why text-only routing signals are insufficient for VLM selection. Routing-relevant information may reside in the image and be absent from the question text. Recent work has begun to incorporate multimodal queries and model profiles into VLM selection \cite{arms2026}. A multimodal query representation makes both the image and question available to the router, while a model profile supplies a candidate-level summary\cite{huang2026vl292929}. Nevertheless, because a single global profile is shared across heterogeneous queries\cite{hu2026adaptive232323}, it may obscure neighborhoods of the multimodal query space in which the same expert succeeds on some inputs but fails on others.

To address this, we propose MHR-Router, a query-adaptive VLM selection framework that estimates expert suitability from multimodal queries and local outcome evidence. Our main contributions are as follows:
\begin{itemize}
    \item We propose MHR-Router, a query-adaptive VLM selection framework that fuses frozen image-text features into a unified multimodal routing representation, overcoming the reliance of prior routers on text-based cues or static model profiles.
    
    \item We introduce a heterogeneous retrieval-enhanced routing mechanism with expert-specific success and failure memories, whose retrieved local evidence enables a shared scorer to capture fine-grained expert competence beyond global profiles.
    
    \item Experiments on 6,400 queries spanning eight visual abilities and eight open-source VLMs show that MHR-Router achieves $71.58{\pm}0.02\%$ routing accuracy, exceeding the development-selected best single expert by 7.36\% and a matched retrieval-free multimodal router by 1.22\%. 
\end{itemize}

\section{Related Work}
\label{sec:related_work}
Cascaded inference \cite{frugalgpt2023}, preference-based
routing \cite{routellm2024}, and in-context model
representations \cite{iclrouter2025} exploit complementary
LLM capabilities. These approaches focus on text-based selection. VLM routing additionally requires visual evidence. CLIP \cite{clip2021} and SigLIP \cite{siglip2023} learn image-text representations, while LLaVA, Qwen-VL, and InternVL support multimodal understanding \cite{qwen2vl2024,qwen25vl2025}. TextVQA \cite{textvqa2019}, CLEVR \cite{clevr2017}, and ScienceQA \cite{scienceqa2022} provide task-specific evaluation, and MVTec AD benchmarks industrial anomaly detection \cite{mvtecad2019}. ARMS studies VLM selection with multimodal queries and VLM profiles, including an expanding model pool \cite{arms2026}. Despite these advances, existing VLM routing methods still struggle to capture expert-specific failure patterns and often rely solely on global model profiles, lacking fine-grained, query-level evidence to distinguish between similarly capable experts.

\section{Method}
\label{sec:method}

\begin{figure}[t]
\centering
\includegraphics[
  width=0.46\textwidth
]{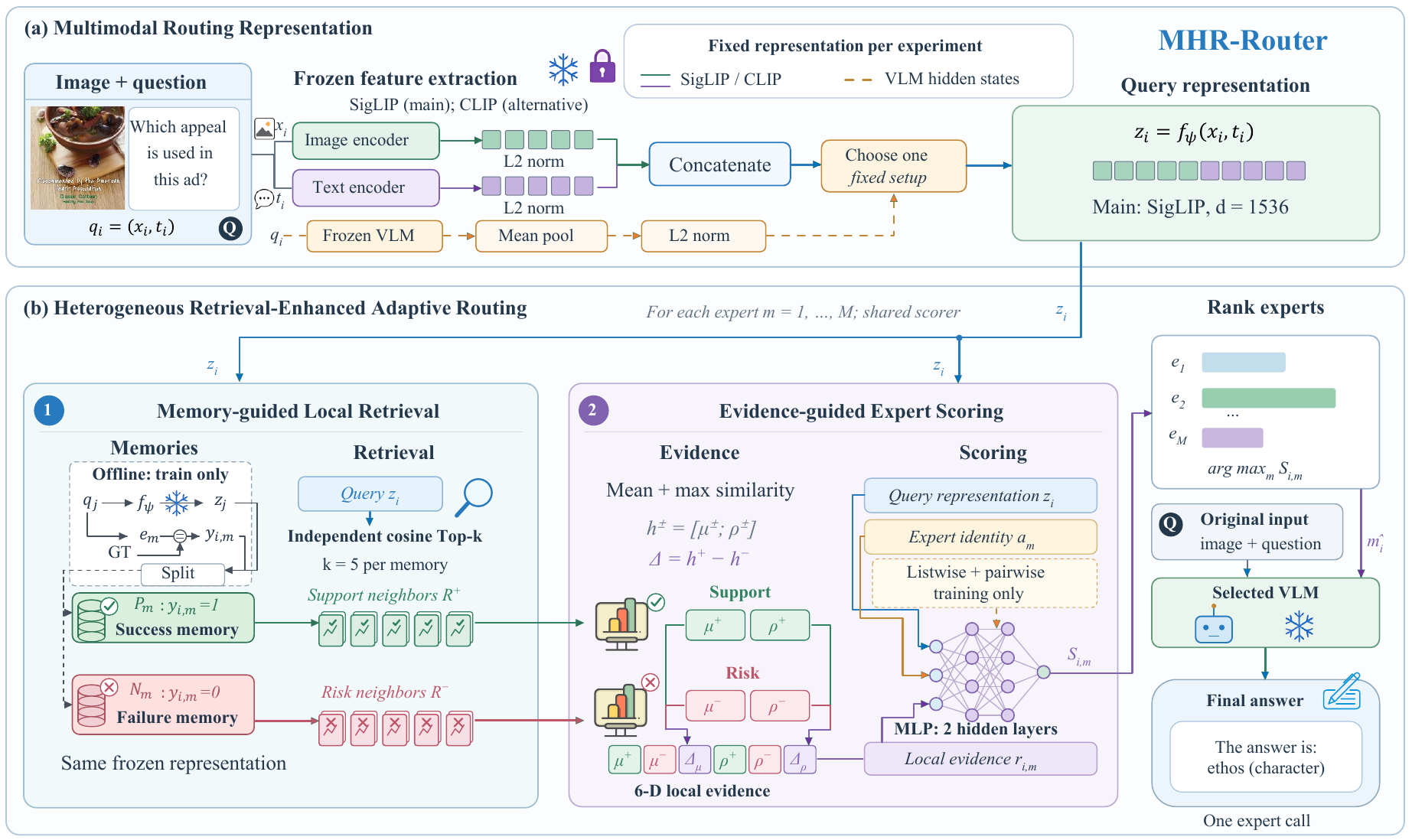}
\caption{Overview of MHR-Router. (a) Frozen multimodal features form the routing representation. (b) Expert-specific success and failure memories provide local retrieval evidence for a shared scorer. The selected expert receives the original query.}
\label{fig:method_overview}
\end{figure}

We propose MHR-Router, a multimodal heterogeneous retrieval-enhanced adaptive routing framework for VLM inference, as illustrated in Fig.~\ref{fig:method_overview}. Its objective is to accurately dispatch each multimodal query to the most suitable VLM expert, leveraging local retrieval evidence to overcome the limitations of static or single-modality routing. MHR-Router consists of two stages: (1) \textit{Multimodal Routing Representation}, which extracts and fuses frozen multimodal features to construct a comprehensive query representation; and (2) \textit{Heterogeneous Retrieval-Enhanced Adaptive Routing}, which retrieves expert-specific success and failure memories to form local evidence and employs a shared scorer to adaptively rank and select the optimal expert for the original query.

\subsection{Multimodal Routing Representation}
\label{subsec:multimodal_representation}

The first stage of MHR-Router constructs a query representation that supports expert ranking without invoking any candidate VLM. This design is motivated by a practical constraint. Routing must be decided before inference, and therefore the router cannot rely on answer-level signals that are only available after all experts have been queried. Instead, the router must operate on a representation that preserves two complementary sources of evidence: the visual content of the image and the intent expressed in the question. 

Let $\mathcal{E}=\{e_1,\ldots,e_M\}$ denote a fixed pool of $M$ VLM experts, and let $q_i=(x_i,t_i)$ denote query $i$, where $x_i$ is the image and $t_i$ is the question. For a training query $q_i$, we define $y_{i,m}\in\{0,1\}$ to indicate whether expert $e_m$ produces a correct answer for $q_i$. Since multiple experts may succeed on the same query, the routing problem is naturally formulated as a ranking problem over experts rather than a single-label classification problem. The router is trained to assign higher scores to correct experts than to incorrect ones, after which one expert is selected for inference. Formally, we encode each query using a frozen multimodal feature extractor $f_\psi$:
\begin{equation}
z_i=f_\psi(x_i,t_i), \qquad z_i\in\mathbb{R}^{d}.
\label{eq:mm_rep}
\end{equation}
where $\psi$ denotes the parameters of the feature extractor and remains frozen throughout router training. Freezing $\psi$ is a deliberate design choice. It prevents the routing representation from drifting toward the answer distributions of specific experts, which would undermine the heterogeneity that the retrieval memories are intended to capture. It also keeps the router lightweight and compatible with a fixed, reusable query index, so that the same representation can be shared by the expert-specific success and failure memories in the second stage.

As shown in Fig.~\ref{fig:method_overview}(a), the main SigLIP configuration constructs $z_i$ by concatenating an $\ell_2$-normalized image embedding and an $\ell_2$-normalized text embedding. The CLIP configuration follows the same construction. In the VLM-hidden-state alternative, final-layer hidden states over valid input tokens are first mean-pooled and then $\ell_2$-normalized. In all cases, the two modalities are normalized before fusion so that neither modality dominates the concatenated representation by scale alone. One representation configuration is fixed for each experiment, and the parameters $\psi$ remain frozen throughout scorer training. This gives a controlled query representation that isolates the effect of heterogeneous retrieval and adaptive routing from changes in the underlying multimodal encoder. The frozen representation $z_i$ is intentionally modality-complete but expert-agnostic.

\subsection{Heterogeneous Retrieval-Enhanced Adaptive Routing}
\label{subsec:history_routing}

The second stage of MHR-Router injects expert heterogeneity into routing through retrieval. The query representation \(z_i\) from Sec.~\ref{subsec:multimodal_representation} is modality-complete but expert-agnostic: it describes the image--question pair without committing to any particular VLM. To decide which expert is most likely to succeed, the router must therefore consult expert-specific evidence. Rather than encoding this evidence in additional learnable expert parameters, we store it non-parametrically as success and failure memories, and retrieve from these memories at inference time. This design keeps the expert pool extensible and allows new experts to be added by populating their memory banks without retraining the shared scorer.

For each expert, ground-truth correctness partitions frozen training representations into success and failure memories:
\begin{equation}
\mathcal{P}_m=\{z_j\mid y_{j,m}=1\}, \qquad
\mathcal{N}_m=\{z_j\mid y_{j,m}=0\}.
\label{eq:history_sets}
\end{equation}
Here \(\mathcal{P}_m\) summarizes the region of the representation space in which expert \(e_m\) has previously succeeded, while \(\mathcal{N}_m\) summarizes the region in which it has previously failed. Together they form a non-parametric profile of the expert's competence relative to the frozen query representation.

Independent cosine top-\(k\) retrieval forms \(\mathcal{R}^{+}_{i,m}\subset\mathcal{P}_m\) and \(\mathcal{R}^{-}_{i,m}\subset\mathcal{N}_m\), providing local support and risk evidence. Training queries exclude themselves from retrieval, so that the evidence for \(q_i\) is always drawn from other queries and never from its own correctness label. Mean and maximum cosine similarity summarize each neighborhood:
\begin{equation}
\begin{aligned}
\mu^{\pm}_{i,m}&=\frac{1}{|\mathcal{R}^{\pm}_{i,m}|}
 \sum_{z_j\in\mathcal{R}^{\pm}_{i,m}}
 \cos(z_i,z_j),\\
\rho^{\pm}_{i,m}&=\max_{z_j\in\mathcal{R}^{\pm}_{i,m}}
 \cos(z_i,z_j),\\
h^{\pm}_{i,m}&=[\mu^{\pm}_{i,m};\rho^{\pm}_{i,m}],\quad
\Delta_{i,m}=h^{+}_{i,m}-h^{-}_{i,m}.
\end{aligned}
\label{eq:local_evidence}
\end{equation}
Both \(h^\pm_{i,m}\) and \(\Delta_{i,m}\) are two-dimensional. The mean similarity \(\mu^{\pm}_{i,m}\) measures the average consistency between the current query and the retrieved neighborhood, whereas the maximum similarity \(\rho^{\pm}_{i,m}\) measures whether at least one highly similar precedent exists. The difference \(\Delta_{i,m}\) then contrasts success evidence against failure evidence for the same expert, which is more informative than either neighborhood alone. The six retrieval statistics are packed as
\begin{equation}
r_{i,m}=[
\mu^+_{i,m};
\mu^-_{i,m};
\Delta_{\mu,i,m};
\rho^+_{i,m};
\rho^-_{i,m};
\Delta_{\rho,i,m}]
\in\mathbb{R}^{6},
\label{eq:retrieval_statistics}
\end{equation}
where
\(\Delta_{\mu,i,m}=\mu^+_{i,m}-\mu^-_{i,m}\) and
\(\Delta_{\rho,i,m}=\rho^+_{i,m}-\rho^-_{i,m}\). With one-hot expert identity \(a_m\in\{0,1\}^{M}\), a shared MLP with two hidden layers scores every candidate:
\begin{equation}
\begin{aligned}
\phi_{i,m}&=[z_i;a_m;r_{i,m}],\\
S_{i,m}&=g_\theta(\phi_{i,m}), \qquad
\hat m_i=\arg\max_{m}S_{i,m}.
\end{aligned}
\label{eq:scoring}
\end{equation}
The scorer parameters are shared across all experts, while \(a_m\) identifies the candidate whose memories provide \(r_{i,m}\). We optimize only scorer parameters \(\theta\). Define correct experts \(\mathcal{C}_i=\{m:y_{i,m}=1\}\) and wrong experts \(\mathcal{W}_i=\{m:y_{i,m}=0\}\). Let \(\mathbf{S}_i\) collect all expert scores and \(p_{i,m}=\operatorname{softmax}(\mathbf{S}_i)_m\). For training queries \(\mathcal{T}\) with both outcomes:
\begin{equation}
\begin{aligned}
\ell_i^{\rm list}
 &=-\frac{1}{|\mathcal{C}_i|}
   \sum_{m\in\mathcal{C}_i}\log p_{i,m},\\
\ell_i^{\rm pair}
 &=\underset{\substack{p\in\mathcal{C}_i\\n\in\mathcal{W}_i}}
   {\operatorname{mean}}
   \operatorname{softplus}(\gamma-S_{i,p}+S_{i,n}),\\
\mathcal{L}
 &=\frac{1}{|\mathcal{T}|}\sum_{i\in\mathcal{T}}
   \left(\ell_i^{\rm list}+\lambda\ell_i^{\rm pair}\right).
\end{aligned}
\label{eq:training_objective}
\end{equation}
The listwise term assigns uniform target mass to all correct experts, while the pairwise term penalizes correct--incorrect score gaps below \(\gamma\). Queries for which all experts are correct or all experts are incorrect provide no ranking supervision and are excluded from scorer optimization. Only \(\theta\) is optimized; the feature extractor and memory banks remain fixed. At inference, \(f_\psi\), the memory banks, and \(g_\theta\) are fixed. For a new query, the router computes \(z_i\), retrieves \(r_{i,m}\) and evaluates \(S_{i,m}\) for every expert, and selects \(\hat m_i=\arg\max_m S_{i,m}\). Only the selected expert receives the original image--question pair. No test correctness label, ability label, source-task label, or global expert-accuracy feature is used for routing.

\section{Experiments}
\label{sec:experiments}

\subsection{Experimental Setup}
\label{subsec:exp_setup}

\noindent\textbf{Benchmark and experts.}
The benchmark contains 6,400 queries drawn from CIFAR-100 \cite{cifar1002009}, Food-101 \cite{food1012014}, CLEVR \cite{clevr2017}, MVTec AD \cite{mvtecad2019}, TextVQA \cite{textvqa2019}, ChartQA \cite{chartqa2022}, ScienceQA \cite{scienceqa2022}, and A-OKVQA \cite{aokvqa2022}, covering eight visual abilities with 800 queries per ability. The image-disjoint train/dev/test split contains 2,064/516/3,820 queries, yielding 30,560 test query--expert pairs. The eight instruction-tuned experts are Qwen2-VL-2B \cite{qwen2vl2024}, Qwen2.5-VL-3B/7B \cite{qwen25vl2025}, LLaVA-OneVision (Qwen2-7B-OV) \cite{llavaonevision2025}, LLaVA-v1.6 (Mistral-7B) \cite{llavanext2024}, Phi-3.5-Vision \cite{phi35vision2024}, and InternVL2.5-2B/4B \cite{internvl252025}. Fixed prompting and decoding apply per answer type. Correctness uses normalized short-answer/yes--no matching, numerical equivalence, multiple-choice extraction, or class matching. For each test query, the router selects exactly one expert. Routing accuracy is the percentage of test queries for which the selected expert's stored response is judged correct.

\noindent\textbf{Implementation details.}
We use $k=5$ per memory and $1536+8+6=1550$ input features. The scorer has two 160-unit hidden layers with GELU, LayerNorm, and dropout 0.45 (274,721 parameters). AdamW \cite{adamw2019} training runs for 80 epochs with learning rate $3\times10^{-4}$, weight decay $5\times10^{-4}$, $\lambda=0.5$, and $\gamma=0.2$ on one RTX 4090. We report the mean $\pm$ sample standard deviation over seeds 13, 42, and 87, selecting each checkpoint by development accuracy.

\begin{figure}[!t]
\centering
\includegraphics[width=\linewidth]{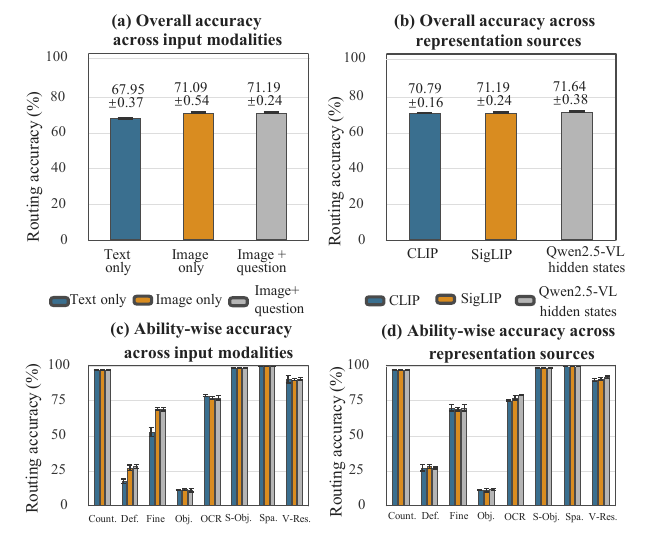}
\caption{Overall and ability-wise routing accuracy across input modalities and representation sources.}
\label{fig:representation_ablation}
\end{figure}

\subsection{Experimental Results}
\label{subsec:exp_results}

\subsubsection{Effect of Multimodal Query Representations}
\label{subsubsec:multimodal_advantage}

Fig.~\ref{fig:representation_ablation}(a,c) compares SigLIP input modalities using the mean and sample standard deviation across three seeds. Routing with image-plus-question features reaches $71.19\%$, compared with $67.95\%$ for text-only features. Image-only features capture most of this gain, while adding the question further improves defect anomaly detection, OCR understanding, and visual reasoning. Figs.~\ref{fig:representation_ablation}(b,d) compares frozen encoders under the image-plus-question setting. The CLIP, SigLIP, and Qwen2.5-VL representations achieve $70.79{\pm}0.16\%$, $71.19{\pm}0.24\%$, and $71.64{\pm}0.38\%$, respectively. Qwen2.5-VL achieves the highest overall mean and higher ability-level means on OCR understanding and visual reasoning, whereas SigLIP is higher on defect anomaly detection. Thus, no encoder dominates all abilities. We adopt SigLIP because its overall accuracy remains close to that of Qwen2.5-VL while keeping the routing representation independent of the candidate expert pool. The larger modality gap indicates that visual evidence contributes more to routing accuracy than the choice among the evaluated encoders, while question information provides complementary ability-specific gains.

\FloatBarrier

\subsubsection{Routing Accuracy}
\label{subsubsec:routing_accuracy}

Table~\ref{tab:main_results} compares fixed and group-level references with learned query-level routers. Best Single applies the expert with the highest overall development accuracy to every test query. The per-ability and per-task rows are diagnostic references that use known group labels. The Multimodal Query Router is the matched retrieval-free baseline: it uses the same query embedding, expert identity, scorer architecture, objective, and training schedule as MHR-Router, but sets all six retrieval statistics to zero. MHR-Router achieves $71.58{\pm}0.02\%$, exceeding Best Single by $7.36\%$ and the Multimodal Query Router by $1.22\%$. Under this matched setup, the latter gap measures the observed gain from adding outcome-specific local retrieval statistics. This demonstrates that MHR-Router effectively leverages evidence retrieved from each expert’s historical successes and failures to improve query-level expert selection.

\begin{table}[!htbp]
\centering
\normalsize
\caption{Main routing accuracy and gains over the development-selected Best Single baseline.}
\label{tab:main_results}
\setlength{\tabcolsep}{2.4pt}
\begin{tabular*}{\linewidth}{@{\extracolsep{\fill}}lcc@{}}
\toprule
Method & Acc. (\%) & Gain (pp) \\
\midrule
Random Router & 51.21 & -13.00 \\
Best Single (dev-selected) & 64.21 & +0.00 \\
Per-ability Dev Selection$^\dagger$ & 67.70 & +3.48 \\
Per-task Dev Selection$^\dagger$ & 67.70 & +3.48 \\
\midrule
Multimodal Query Router & $70.36{\pm}0.38$ & +6.14 \\
MHR-Router (Ours) & $\mathbf{71.58{\pm}0.02}$ & $\mathbf{+7.36}$ \\
\bottomrule
\end{tabular*}
\par
\vspace{1pt}
\parbox{\linewidth}{\normalsize $^\dagger$Uses development performance and the known group label only as a diagnostic reference.}
\end{table}

\begin{table}[!htbp]
\centering
\normalsize
\caption{Ability-wise routing accuracy relative to the development-selected expert for each ability.}
\label{tab:ability_results}
\setlength{\tabcolsep}{1.2pt}
\begin{tabular*}{\linewidth}{@{\extracolsep{\fill}}lcccc@{}}
\toprule
Ability & $N$ & MHR-Router & Dev-best & Gain (pp) \\
\midrule
Counting & 480 & $96.88{\pm}0.00$ & 96.88 & +0.00 \\
Defect anomaly & 480 & $27.71{\pm}0.75$ & 19.17 & +8.54 \\
Fine-grained recog. & 480 & $72.43{\pm}0.43$ & 50.83 & +21.60 \\
Object recog. & 480 & $11.39{\pm}0.32$ & 11.04 & +0.35 \\
OCR understanding & 467 & $77.02{\pm}2.55$ & 79.44 & -2.43 \\
Small-object detection & 480 & $98.33{\pm}0.00$ & 97.71 & +0.62 \\
Spatial reasoning & 480 & $99.38{\pm}0.00$ & 99.58 & -0.21 \\
Visual reasoning & 473 & $89.92{\pm}1.41$ & 87.53 & +2.40 \\
\bottomrule
\end{tabular*}
\par
\end{table}

Table~\ref{tab:ability_results} reports ability-wise routing accuracy on the test set relative to the development-selected expert for each ability, where $N$ denotes the number of test queries. MHR-Router improves five of eight abilities, matches counting, and achieves its largest gains in fine-grained recognition (+21.60 pp) and defect anomaly detection (+8.54 pp). These results indicate that MHR-Router can effectively leverage query-level variations to improve routing performance. This suggests that retrieval memories are most useful when global expert accuracy is a poor proxy for per-query competence.

\FloatBarrier

\subsubsection{Parameter Sensitivity Analysis}
\label{subsubsec:parameter_sensitivity}
\begin{figure}[!htbp]
\centering
\includegraphics[width=\linewidth]{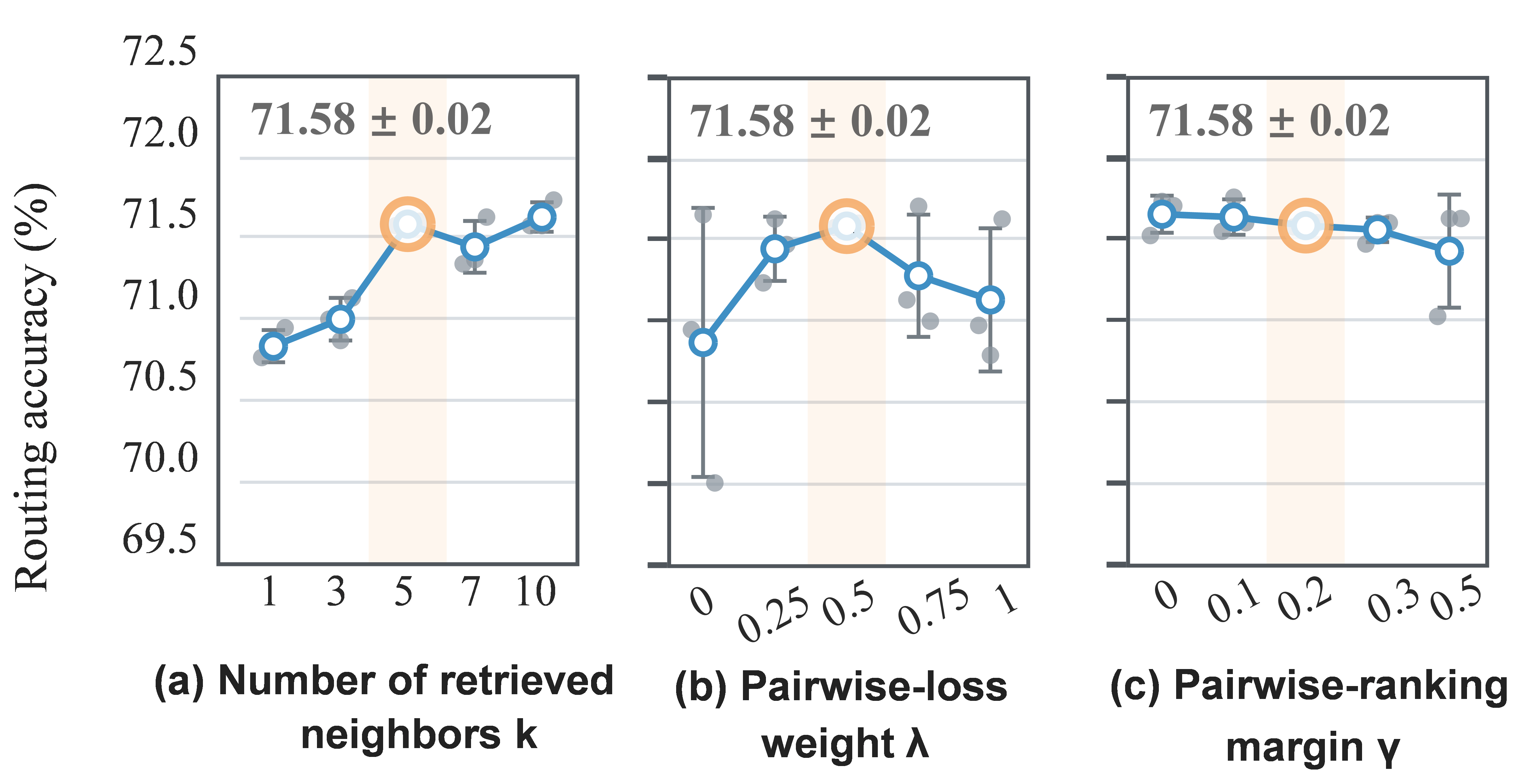}
\caption{Parameter sensitivity analysis of MHR-Router.}
\label{fig:parameter_sensitivity}
\end{figure}

Fig.~\ref{fig:parameter_sensitivity} varies one hyperparameter at a time over three seeds under the same inference setting as MHR-Router, without ability labels, source-task labels, or global expert-accuracy features. Accuracy rises from $70.83{\pm}0.10\%$ at $k=1$ to $71.58{\pm}0.02\%$ at $k=5$ and changes by at most $0.14\%$ thereafter. For the loss weight, $\lambda=0.5$ yields the highest mean and the lowest variation, whereas removing the pairwise term yields $70.86{\pm}0.83\%$. Mean accuracy varies by only $0.23\%$ across the tested values of $\gamma$. These results support $k=5$, $\lambda=0.5$, and $\gamma=0.2$ as stable defaults and confirm that routing performance is driven by retrieved query-level evidence rather than by per-group hyperparameter tuning. This is consistent with the design goal of MHR-Router. Routing decisions should be driven by retrieved query-level evidence, not by manually tuned per-group settings.

\subsection{Ablation Study}
\label{subsec:ablation}

We retrain three matched feature-group ablations using the same data split, frozen SigLIP representations, seeds, 1,550-dimensional scorer input, architecture, and optimization protocol as the full model. To preserve input size and scorer capacity, the ablation without positive evidence sets the positive-memory mean and maximum cosine similarities and both positive--negative difference features to zero. The ablation without negative evidence analogously sets the negative-memory mean and maximum cosine similarities and the same difference features to zero. MHR-Router achieves $71.58{\pm}0.02\%$, compared with $70.41{\pm}0.54\%$ without positive evidence, $70.34{\pm}0.28\%$ without negative evidence, and $70.36{\pm}0.38\%$ without retrieval evidence. These correspond to decreases of $1.17\%$, $1.24\%$, and $1.22\%$, respectively. The full model exceeds every ablation in all three seeds, supporting the joint use of success, failure, and contrast evidence.

\section{Conclusion}
\label{sec:conclusion}

This paper presents MHR-Router, a retrieval-augmented framework for adaptive VLM selection that combines frozen multimodal representations with expert-specific success and failure memories. By leveraging complementary evidence from successful and failed cases, MHR-Router enables input-dependent expert selection without requiring ability or source-task labels at inference time. Experiments across eight abilities and eight experts demonstrate that MHR-Router outperforms the best single expert selected on the development set by 7.36\% and the matched retrieval-free Multimodal Query Router by 1.22\%.

\clearpage
\bibliographystyle{IEEEbib}
\bibliography{strings,refs}

\end{document}